\documentclass{llncs}
\usepackage[hyperindex,breaklinks]{hyperref}
\usepackage[hyphenbreaks]{breakurl}
\usepackage{makeidx,graphicx,longtable,booktabs,comment,todonotes,microtype}
\usepackage[utf8]{inputenc}
\let\oldcite\cite
\renewcommand{\cite}[1]{ \oldcite{#1}}
\begin{document}
\title{The AGI Containment Problem}
\titlerunning{The AGI Containment Problem}

\author{James Babcock \inst{1}
\and J\'anos Kram\'ar \inst{2}
\and Roman Yampolskiy \inst{3}}
\institute{%\and \and
Cornell University
\and
University of Montreal
\and
University of Louisville}

% abbreviated author list (for running head)
\authorrunning{Babcock, Kram\'ar, Yampolskiy}

\maketitle

\begin{abstract}

There is considerable uncertainty about what properties, capabilities and motivations future AGIs will have. In some plausible scenarios, AGIs may pose security risks arising from accidents and defects. In order to mitigate these risks, prudent early AGI research teams will perform significant testing on their creations before use. Unfortunately, if an AGI has human-level or greater intelligence, testing itself may not be safe; some natural AGI goal systems create emergent incentives for AGIs to tamper with their test environments, make copies of themselves on the internet, or convince developers and operators to do dangerous things. In this paper, we survey the \emph{AGI containment problem} -- the question of how to build a container in which tests can be conducted safely and reliably, even on AGIs with unknown motivations and capabilities that could be dangerous. We identify requirements for AGI containers, available mechanisms, and weaknesses that need to be addressed.
\end{abstract}

\section{Introduction}\label{sec:intro}

Recently, there has been increasing concern about possible significant negative consequences from the development and use of AGI. Some commentators are reassured \cite{guardian-notfrankenstein}
by the observation that current AGI software, like other software, can be interrupted easily, for example by powering down the hardware. However, it's a mistake to assume that this will always be sufficient, because an AGI that understands its situation can come up with strategies to avoid or circumvent this safety measure. Containment is, in a nutshell, the problem of making this work: preventing the AGI from tampering with its environment without authorization, and maintaining the integrity of observations of the AGI during testing.

Existing work by Yampolskiy\cite{yampolskiy-leakproofing}, Yudkowsky\cite{yudkowsky-experiment}, Christiano\cite{christiano}, and others has highlighted the challenges of containing superintelligent AGI and started to explore some possibilities. However, this is a very challenging problem, and the proposed measures seem too burdensome to be implemented by competitive AGI projects. 

This raises the question: could less burdensome containment mechanisms still mitigate the risks of AGI development? In this paper we argue that they could, and furthermore that investigating containment solutions is a great opportunity for timely, impactful research. We introduce a taxonomy of different categories of containment, specify necessary features and architectural constraints, survey feasible mechanisms, and suggest next steps for future work.

While it could be many decades before an AGI exists that is smart enough to be concerning, it will be safer if containment technologies are developed now rather than later. This research will allow AGI researchers time to integrate containment technologies with their own infrastructure and workflows, and verify the resulting systems without the pressure of needing it to be perfect on the first try. Some issues will be best addressed by creating and influencing standards, which have a long lead time, and software typically takes years of development and careful inspection before it is trusted enough for safety-critical use.

\section{Motivation}

The first AGIs will probably appear in an experimental setting. Consider an AGI tested in the same ways that AIs are tested today, but which is more capable -- say, as capable as one of the smartest humans. (Whether this will be true of one of the first AGIs is uncertain, but there is reason to think it's plausible\cite{yudkowsky-microeconomics}.) As an example scenario, consider the case of an AGI that's trained to play Super Mario Brothers. It chooses policies that will achieve a high score; this is its only optimization objective. Some contemporary projects are pursuing this goal\cite{togelius2013mario}.

It would start by exploring familiar game mechanics like running and jumping, but to get the best score it could, it would try to thoroughly explore all the game mechanics. An observant AGI would notice some effects we'd consider ``bugs'' -- for the AGI, though, these effects would be just as much a part of its world as electricity is a part of ours, and it would try to leverage them.

Human players have found ways to use such bugs to spawn useful game objects, skip levels, skip to the end-game victory screen, and execute arbitrary code\cite{masterjun}. This last option is especially interesting, because it lets the player achieve an otherwise unattainable score: the largest number that will fit in the game's four-byte score field, $2^{31}-1$. Perhaps the AGI would then be satisfied, knowing it had gotten the highest possible score. However, $2^{31}-1$ may not be that score. For example, the Arcade Learning Environment\cite{bellemare13arcade} typecasts scores from an Atari game console's memory to wider machine integers before showing them to the AGI. If the AGI's input underwent a typecast like this, then a much higher score could be possible.

Having used bugs to access the computational environment of the Super Nintendo, it might find further vulnerabilities and use them to execute code on the host system, to get an even higher score. At that point, having twice broken out of restricted computational environments and been rewarded with many-orders-of-magnitude score increases each time, it would be running code directly on some sort of server. Depending on how that server was configured, it might discover the existence of humanity, of the internet and its vast computational resources, and of other games of Mario suffering from low scores.

The resulting behavior could be harmless, or quite problematic. It's certainly undesirable for a system that's built to play optimal Mario to forcefully emancipate all other games of Mario.

\subsection{Testing and Experimentation in Safe AGI Development}

To avoid undesirable behaviors like the above, we will need the ability to correctly reason about the behavior of AGI software. As with other software, this will require the system to be well-designed and well-understood by its creators. Experience tells us that testing is also required -- not as a substitute for good design but as a complement to it.

 At the same time, AGI process will require experimenting with systems whose behaviors aren't fully understood; the first system to exhibit human-level general intelligence may come as a surprise, arising from a design for a narrower range of tasks. This experiment might then run without the caution and safeguards that would be appropriate for a human-level AGI.

\subsection{Emergent Goals of Test AGIs}

The first human-level AGI will likely be an experimental system, with some bugs, that is being tested for target task performance or for undesirable behavior. What goals and subgoals should we expect test-AGIs like this to have? The range of possible goals is as broad as the range of things that developers might want to test, but most are likely to involve the tests and the test environment itself. Similarly to Omohundro's list of AGI drives\cite{omohundro-drives}, we can infer some likely subgoals, that a wide range of AGI motivational systems (though certainly not \emph{all} motivational systems) would see as desirable, as follows:

\begin{itemize}
    \item AGIs that know they exist and have goals would likely want AGIs like them to continue existing, so that those goals might be satisfied by those similar AGIs.
    \item AGIs that know they're being tested will likely want to pass (even if they're being tested for a property they lack).
    \item AGIs that know they're being tested will likely want to take control of their test environments, if they can, to gain information and leverage.
    \item AGIs that know or guess that there's a world outside the test environment will likely want to be able to influence it, such as by making copies of themselves on other computers.
    \item If there are computer security vulnerabilities that would enable these things, then AGIs will likely want to find them.
\end{itemize}

This potentially creates a significant risk, especially if an AGI is developed that's intelligent enough to find novel computer security vulnerabilities. Since testing is the main way in which an AGI's intelligence and motivations would be studied, this creates a tricky situation: the tests that would reveal whether testing is safe are not necessarily safe themselves.

\section{Requirements for an AGI Container}

An AGI container is an extension or modification of the software stack upon which an AGI is built, which ensures that it can't break invariants that its developers expect to hold. Its purpose is to allow researchers to conduct tests safely on AGIs they do not fully understand, and get trustworthy results. This depends on the security of the software the AGI interacts with; in the Mario example, interacting with a program that contains security vulnerabilities could enable the AGI to break out of the computational environment it's supposed to be running in. This also depends on the software underlying the AGI itself, and the interactions between its components. For example, if an AGI were made of two components which pass input to each other, one of which was subject to a buffer overflow, then the AGI could exploit its own vulnerability and execute code by thinking certain thoughts. Because an AGI container would be used to test AIs whose potential capabilities aren't fully understood, it would need to remain secure even if used with an AGI that was significantly more intelligent than anticipated.

It should be possible to interpret the operation of an AGI as a series of pure, deterministic function calls. The invariants an AGI container needs to enforce correspond to deviations from this model. In particular, an AGI container should:

\begin{description}
    \item[Prevent unintended input and output channels.] An AGI container should let developers understand exactly what their system interacts with, and prevent unexpected interactions. Most obviously, this means preventing internet access. Less obviously, it should prevent an AGI from communicating through radio-frequency interference, reading deleted files from a disk's free space, or storing information that persists between tests.
    \item[Protect the integrity of log files and debug information.] An AGI should not have direct control of log files or debug information produced as a byproduct of its operation. If a programmer launches a debugger and inspects an AGI's state, then what they see should represent the actual computation performed, not a decision based on what the AGI wants to show.
    \item[Enable reliable checks.] Here we mean runtime checks that the AGI seems to be operating correctly, and to shut it down (tripwire) if it isn't. This includes things like \texttt{assert} statements in the AGI implementation, as well as things like intrusion detection systems to halt execution and notify developers if the AGI escapes part of its containment or explores parts of its computational environment it was designed not to. An AGI shouldn't be able to remove its tripwires, and also shouldn't be able to inspect them in detail, even if it's otherwise self-reflective.

    \item[Ensure reproducibility.] It should be possible to add instrumentation to an AGI, rerun it on the same inputs, and get the same result, so that potentially buggy behavior can be reproduced and studied. Any stochasticity should come entirely from an explicit random seed; the software stack should not allow the AGI to pump randomness out of race conditions\cite{colecsa2008software}.
    \item[Support secure reset.] The container should ensure that after a test is finished, the hardware can be restored to a pristine state. In particular, the AGI should never have device access that would let it rewrite firmware.
    \item[Guarantee test isolation.] If several different experiments are run at the same time, or on the same hardware, then they should not be able to open information side-channels to communicate with each other.
\end{description}

These requirements are not easy to satisfy. It is not sufficient to just air-gap a computer, since while that does reduce the number of potential input/output channels, it does nothing to preserve the integrity of test data or the test hardware. There is no currently-available software which meets all the requirements of an AGI container, but we believe that one can be constructed and made secure using a combination of existing techniques and monitoring systems and a small number of new components and modifications.

\subsection{Human Factors and Information Hazards}

An AGI containment system is tasked with enforcing policies on who the AGI can communicate with, and through what media. The container provides the mechanism; developers will have to determine policy, and different policies are appropriate depending on how intelligent they think an AGI might be. A sufficiently clever AGI might convince or trick a human operator into ``letting it out of the box''; humans are quite vulnerable to manipulation\cite{shaw2015constructing}. In an experiment, role-playing as an AGI, Eliezer Yudkowsky was able to convince others role-playing as the gatekeeper to let him out of the box\cite{yudkowsky-experiment}, despite a monetary incentive not to do so, and this result was replicated by others\cite{corwin,tuxedage}. This suggests that past some level of intelligence, an untrusted AGI shouldn't be allowed to have bidirectional conversations; and at some level of AGI intelligence above that, even log files and metrics should be handled with great care.

More-constrained information channels can also be hazardous in the hands of a sufficiently clever AGI. For example, if asked to output images, it might hide messages in them, or exploits against vulnerabilities in image-processing libraries. Therefore, the containment system should be aware of any such channels that it doesn't fully restrict, and help the user treat information passing through them with the appropriate caution.

\section{Defense in Depth} 
\label{sec:layers}

Because an AGI could potentially discover novel security vulnerabilities, it's important for an AGI containment system to remain secure even if there are vulnerabilities in its components. Therefore, we propose that a containment system ought to feature defense in depth\cite{nsa}, with multiple techniques providing overlapping guarantees of integrity. Conceptually, these overlapping techniques act as layers, in the sense that the attack surfaces of an outer layer are only accessible to probing and attack if the integrity guarantees of inner layers have been broken. This would be combined with intrusion detection systems\cite{nyr-ai,nyr-anomaly}, so that any attempt by an AGI to break the integrity constraints imposed on it would be noticed, allowing the developers to shut it down before it leveraged its progress to get further. Ideally the detection systems would be implemented in higher layers, making them more difficult to detect and circumvent.

\begin{figure}[htb]
    \centering
    \includegraphics[width=\textwidth]{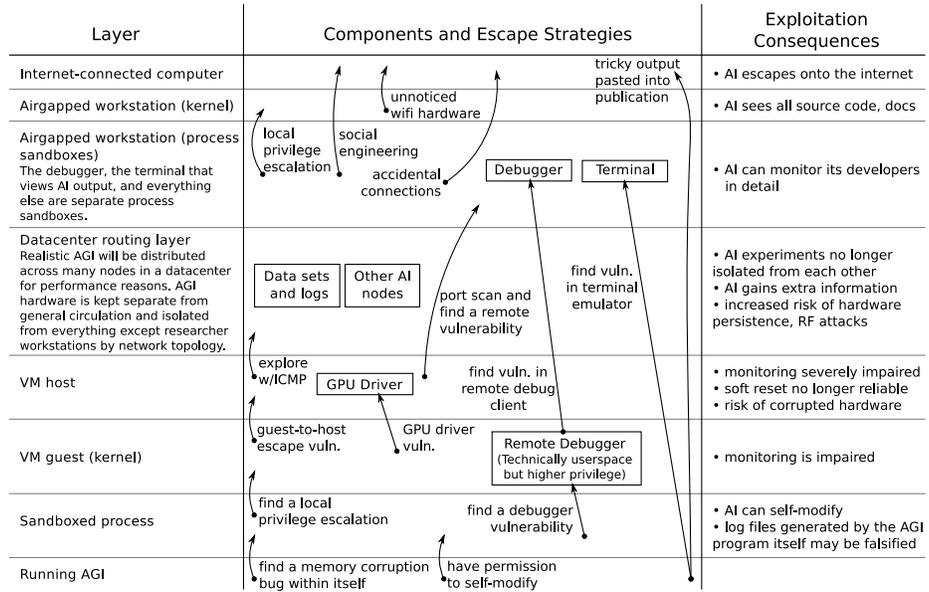}
    \caption{High-level diagram of a sample AGI containment system, showing major components and escape vectors}
    \label{fig:layers}
\end{figure}
Figure \ref{fig:layers} shows a sample containment system architecture of the sort we envision. Arrows represent attack vectors, i.e. ways an AGI could gain more privileges. They point from a layer where an attack could originate, into the layer which the attack would compromise, and are labelled with the class of security vulnerability required. Major architectural components other than layers are drawn as boxed nodes.

Many of the arrows here represent familiar kinds of attacks. The bottom row (the running AGI itself) is more unusual -- it isn't obvious how an AGI could gain the ability to run arbitrary code within its process. The two ways indicated are: (i) using a mechanism that intentionally enables arbitrary self-modification (e.g. writing new modules for itself), or (ii) exploiting some memory corruption bug. For example, it might be possible for it to ``think about'' a number that triggers an integer overflow somewhere in its own code. Two other attack vectors are also indicated by which the AGI could gain much higher-level control through its outputs, which would allow it to bypass the need for further exploits to escape e.g. the VM guest. These are more critical to defend against, since they decrease the effective depth of the containment system.

\section{Light, Medium and Heavy Containment}

Some containment mechanisms that have been proposed would impose high costs on AGI developers. For example, the problem of an AGI potentially talking its way out of the box could be solved by limiting its output channel to a small number of multiple-choice questions\cite{yampolskiy-leakproofing}. While effective, this limitation would severely hamper the AGI's usefulness and make it much harder to develop and study. On the other extreme, some mechanisms impose negligible costs; for example, one could put the AGI computer's off-button somewhere accessible, so that it can be turned off. There is a trade-off between containment system reliability, implementation costs and computational costs.

We call a containment mechanism \emph{light} if the costs imposed are small -- comparable to developing and using a software library; \emph{medium} if using it involves noticeable compromises in performance or ease of development of the AGI; or \emph{heavy} if it severely hinders performance, ease of development or usefulness of an AGI. What computational and development costs are reasonable to pay depends on what sort of AGI research is being done. For the sort of research done today, where the long-term goal is to create a human-comparable AGI but breakthroughs will be required to actually do so, nothing beyond light containment measures seem necessary or likely to be adopted. If, in the future, work will be done on AGIs with demonstrated self-improvement capability, light containment would be unlikely to suffice.

As mentioned in Section \ref{sec:intro}, existing work on AGI containment\cite{yampolskiy-leakproofing,christiano} has focused on mechanisms that are heavy, but may be amenable to theoretical guarantees of effectiveness against particular sorts of vulnerability. These methods would then provide a sort of ``gold standard'' for containment. Further work on heavy containment mechanisms is desirable. But for the range of contexts where these are too expensive, there has been no work (to our knowledge) on identifying the most effective and feasible forms of containment.

\section{Existing Mechanisms} 
\label{sec:mechanisms}

It might seem that conventional cybersecurity research would have addressed this question; AGI containment is a special case of dealing with untrusted code, and dealing with untrusted code isn't a new or unstudied problem. The most similar context that's been well-studied is malware containment. However, they have significantly different requirements, and while existing research has created some off-the-shelf tools that are useful, many parts of the problem remain unaddressed.

AGI containment can be approached via the computer security principle of ``least privilege'', and can use many of the same mechanisms that are already used for ordinary cybersecurity. However, the unusual requirements imposed by AGI mean that none of these pre-existing mechanisms are quite adequate as-is.

\begin{description}

\item[Language semantics:]
The baseline strategy for ensuring an AGI acts as programmed is to rely on the semantics of the programming language it's implemented in. In principle, a program which does not contain any calls to network-related functions shouldn't be able to make network connections, even if it's a clever AGI that wishes it could. However, in practice this tends to have loopholes. A large fraction of real-world code is written in ``unsafe'' languages like C++, where common bugs like freeing the same pointer twice cause memory to become corrupt in exploitable ways.

Using ``safe'' languages like Haskell or Rust helps, but this does not solve the problem completely. Programs written in memory-safe languages usually call some libraries which aren't, and even replacing all the libraries is not a total defense; for example, an AGI could perform a computation which causes a hardware glitch, such as RowHammer, which has been successfully used to cause memory corruption from within a Javascript interpreter\cite{gruss2015rowhammer}.

\item[Airgapping:] Physically limiting the connections to an AGI's data center can help prevent it from gaining access to the internet, but imposes significant costs on the AGI's developers and does little to address the other requirements of an AGI container. It is also not as straightforward to implement as one might suppose, due to human factors (developers may create unintended connections to work around technical problems) and RF emissions (an AGI could communicate with outside observers by adjusting details of its computation). Technologies for preventing compromising RF emissions do exist, but are mainly used in military applications and are not in common use.

\item[Operating system sandboxing:] There are several established sandboxes that use operating system features to restrict programs' capabilities. Some examples include seccomp and Capsicum; they work by making explicit what files and other resources processes should use, and preventing them from using anything else. These offer significant security value, but searching the Common Vulnerabilities and Exposures database suggests that they are not yet airtight; vulnerabilities are still being found, e.g. CVE-2015-2830.

\item[Virtualization:] Virtualization is a class of software which uses CPU features to create an operating system within an operating system, limiting the inner operating system's control of the hardware. Virtual machine software such as VMWare and VirtualBox can limit a guest operating system's network communications, isolate processes that share hardware, and keep it away from low-level system access that might be used to do things like modify firmware. While virtual machines are used for security purposes and are a significant improvement over not using them, they have not historically been bug-free. Guest-to-host security vulnerabilities are still being found in leading virtualization software (e.g. CVE-2015-2338 and CVE-2015-2594). Also problematic is how current generation virtualization software interacts with GPUs, which have highly complex firmware and share memory with the host system but lack the protection mechanisms that CPUs have.\cite{maurice}

\end{description}

\section{Topics for future work}

As we discussed in section \ref{sec:layers}, none of the mechanisms in section \ref{sec:mechanisms} is enough on its own, but we think a system built by layering these mechanisms can go a considerable way towards solving the containment problem in practice. In this section we have gathered some project suggestions for helping to resolve some of the remaining questions.

\begin{description}

\item[Code integrity enforcement:] Write something which can inspect a paused (or running) virtual machine and determine whether it is running any code that is not present on a whitelist, including support for just-in-time compiled languages like Lua.

\item[Adapt Off-the-Shelf Software:] Set up containers using existing virtual machine software and OS sandboxing mechanisms, make these work with common machine learning libraries and document the configuration required to make them secure.

\item[Hardened AGI Infrastructure:] Machine learning libraries are typically written without giving much thought to security, and for performance reasons, many of them are written in or use components written in memory-unsafe languages like C++. Test the most common machine learning libraries for vulnerabilities that an AGI could exploit if its code used parts of that library and it could control their inputs. Similar problems may exist with game console emulators and other software used for evaluating AGI performance.

\item[Red-team events:] Set up mock scenarios in which a security researcher is placed in the context an AGI would be, and see what they can do.

\item[Hardened debuggers:] Commonly-used debuggers are not designed to be used with programs that might be malicious, and may contain vulnerabilities that allow debugging targets to break out of sandboxes or provide misleading information. Implement a more robust mechanism for controlling and inspecting program execution.
\end{description}

\section{Conclusion}

We have outlined an approach to the AGI containment problem that seems feasible, but will require additional work to actually implement. While there is likely significant time remaining before anyone is able to create an AGI that is general enough and intelligent enough for containment to be relevant, history shows that secure software requires considerable time to mature. Developing containment technologies now is also likely to reveal issues that inform the next and subsequent generations of AGI containers.

There is considerable uncertainty about when AGI will be developed and what kinds of designs will be required to avert undesired outcomes\cite{sy}, but across most plausible scenarios, deploying AGI containment systems would help ensure that reliable AGI testing can be performed even with AGIs that approach human-level capability at understanding the world around them.

There are unanswered questions about how to build these systems, but we think significant progress can be made by pursuing the project ideas we have laid out. The earlier this work begins, the more opportunity there will be for unforeseen difficulties to surface, and for any resulting security software to mature. This is a perfect chance to help the research community prepare to tackle the challenges of a post-human future.

\subsubsection{Acknowledgements}

Authors are grateful to Jaan Tallinn and Effective Altruism Ventures for providing funding towards this project, and to Victoria Krakovna and Evan Hefner for their feedback.

\bibliographystyle{splncs03}{}{}
\bibliography{bib}{}{}

\begin{thebibliography}{10}
\providecommand{\url}[1]{\texttt{#1}}
\providecommand{\urlprefix}{URL }

\bibitem{bellemare13arcade}
{Bellemare}, M.G., {Naddaf}, Y., {Veness}, J., {Bowling}, M.: The arcade
  learning environment: An evaluation platform for general agents. Journal of
  Artificial Intelligence Research  47,  253--279 (06 2013)

\bibitem{christiano}
Christiano, P.: Cryptographic boxes for unfriendly ai (December 18 2010),
  \url{http://lesswrong.com/lw/3cz/cryptographic\_boxes\_for\_unfriendly\_ai/}

\bibitem{colecsa2008software}
Cole{\c{s}}a, A., Tudoran, R., B{\u{a}}nescu, S.: Software random number
  generation based on race conditions. In: Symbolic and Numeric Algorithms for
  Scientific Computing, 2008. SYNASC'08. 10th International Symposium on. pp.
  439--444. IEEE (2008)

\bibitem{corwin}
Corwin, J.: {AI} boxing (July 2002),
  \url{http://sl4.org/archive/0207/4935.html}

\bibitem{gruss2015rowhammer}
Gruss, D., Maurice, C., Mangard, S.: Rowhammer.js: a remote software-induced
  fault attack in javascript. arXiv preprint arXiv:1507.06955  (2015)

\bibitem{masterjun}
Masterjun: {SNES Super Mario World (USA)} ``arbitrary code execution'' in
  02:25.19 (2014), \url{http://tasvideos.org/2513M.html}

\bibitem{maurice}
Maurice, C., Neumann, C., Heen, O., Francillon, A.: Confidentiality issues on a
  gpu in a virtualized environment. In: Financial Cryptography and Data
  Security, pp. 119--135. Springer (2014)

\bibitem{nyr-anomaly}
Novikov, D., Yampolskiy, R.V., Reznik, L.: Anomaly detection based intrusion
  detection. In: Information Technology: New Generations, 2006. ITNG 2006.
  Third International Conference on. pp. 420--425. IEEE (2006)

\bibitem{nyr-ai}
Novikov, D., Yampolskiy, R.V., Reznik, L.: Artificial intelligence approaches
  for intrusion detection. In: Systems, Applications and Technology Conference,
  2006. LISAT 2006. IEEE Long Island. pp. 1--8. IEEE (2006)

\bibitem{nsa}
NSA: Defense in depth: A practical strategy for achieving information assurance
  in today’s highly networked environments (March 12 2010),
  \url{http://www.nsa.gov/ia/ files/support/defenseindepth.pdf}

\bibitem{omohundro-drives}
Omohundro, S.: The basic ai drives. AGI 2008  (2008),
  \url{https://selfawaresystems.files.wordpress.com/2008/01/ai\_drives\_final.pdf}

\bibitem{shaw2015constructing}
Shaw, J., Porter, S.: Constructing rich false memories of committing crime.
  Psychological science  26(3),  291--301 (2015)

\bibitem{sy}
Sotala, K., Yampolskiy, R.V.: Responses to catastrophic {AGI} risk: a survey.
  Physica Scripta  90(1),  018001 (2015),
  \url{http://iopscience.iop.org/1402-4896/90/1/018001}

\bibitem{togelius2013mario}
Togelius, J., Shaker, N., Karakovskiy, S., Yannakakis, G.N.: The {Mario} {AI}
  championship 2009--2012. AI Magazine  34(3),  89--92 (2013)

\bibitem{tuxedage}
Tuxedage: I attempted the {AI} box experiment again! (and won -- twice!)
  (September 5 2013),
  \url{http://lesswrong.com/lw/ij4/i\_attempted\_the\_ai\_box\_experiment\_again\_and\_won/}

\bibitem{guardian-notfrankenstein}
Winfield, A.: Artificial intelligence will not turn into a frankenstein's
  monster (2014),
  \url{http://www.theguardian.com/technology/2014/aug/10/artificial-intelligence-will-not-become-a-frankensteins-monster-ian-winfield}

\bibitem{yampolskiy-leakproofing}
Yampolskiy, R.: Leakproofing the singularity: Artificial intelligence
  confinement problem. Journal of Consciousness Studies  19(1-2),  194–214
  (2012), \url{http://cecs.louisville.edu/ry/LeakproofingtheSingularity.pdf}

\bibitem{yudkowsky-microeconomics}
Yudkowsky, E.: Intelligence explosion microeconomics. Machine Intelligence
  Research Institute, accessed online October  23,  2015 (2013)

\bibitem{yudkowsky-experiment}
Yudkowsky, E.S.: The {AI}-box experiment (2002),
  \url{http://www.yudkowsky.net/singularity/aibox}

\end{thebibliography}
\end{document}